\documentclass[wcp]{jmlr}

\usepackage{multirow}
\usepackage{bbding}

\def\ie{\emph{ie.}\xspace}

\usepackage{longtable}

\usepackage{booktabs}

\usepackage{lineno}

\makeatletter
\let\Ginclude@graphics\@org@Ginclude@graphics 
\makeatother

\jmlrvolume{222}
\jmlryear{2023}
\jmlrworkshop{ACML 2023}

\title[The Importance of Anti-Aliasing in Tiny Object Detection]{The Importance of Anti-Aliasing in Tiny Object Detection}

\author{\Name{Jinlai Ning} \Email{jinlai.ning@kcl.ac.uk}\\
\Name{Michael Spratling} \Email{michael.spratling@kcl.ac.uk}\\ \addr King's College London, Department of Informatics, London. UK.}

\editors{Berrin Yan{\i}ko\u{g}lu and Wray Buntine}

\begin{document}
\maketitle
\begin{abstract}
Tiny object detection has gained considerable attention in the research community owing to the frequent occurrence of tiny objects in numerous critical real-world scenarios. However, convolutional neural networks (CNNs) used as the backbone for object detection architectures typically neglect Nyquist's sampling theorem during down-sampling operations, resulting in aliasing and degraded performance. This is likely to be a particular issue for tiny objects that occupy very few pixels and therefore have high spatial frequency features. This paper applied an existing approach WaveCNet for anti-aliasing to tiny object detection. WaveCNet addresses aliasing by replacing standard down-sampling processes in CNNs with Wavelet Pooling (WaveletPool) layers, effectively suppressing aliasing. We modify the original WaveCNet to apply WaveletPool in a consistent way in both pathways of the residual blocks in ResNets. Additionally, we also propose a bottom-heavy version of the backbone, which further improves the performance of tiny object detection while also reducing the required number of parameters by almost half. Experimental results on the TinyPerson, WiderFace, and DOTA datasets demonstrate the importance of anti-aliasing in tiny object detection and the effectiveness of the proposed method which achieves new state-of-the-art results on all three datasets. Codes and experiment results are released at \url{https://github.com/freshn/Anti-aliasing-Tiny-Object-Detection.git}.
\end{abstract}

\section{Introduction}
Tiny object detection is a specialized area within the object detection field that focuses on identifying and localising small objects in images. While these objects are prevalent in real-world scenarios such as maritime rescue, driving assistance and security monitoring, they are difficult for standard object detection methods~\citep{faster, cascade, ssd, retinanet, fcos, deform_detr} because of the limited spatial extent and low resolution of the target objects. The key distinction between data used in tiny object detection, such as the TinyPerson~\citep{tinyperson}, WiderFace~\citep{widerface} and DOTA~\citep{dota} datasets, and general object detection datasets lies in the proportion of small and tiny objects: ones that occupy only a few pixels in an image (typically less than 20x20 pixels in TinyPerson).

Due to their small size, the distinguishing features of tiny objects tend to have high spatial frequency. However, such high spatial frequencies will tend to be filtered out by the down-sampling performed within the Convolutional Neural Networks (CNNs) that are typically used as the feature-extracting backbones of object detectors. Furthermore, the methods used for downsampling in these CNNs ignore the issue of aliasing. 

Aliasing refers to the phenomenon where high-frequency information in an image is incorrectly represented or distorted during down-sampling~\citep{dip}. According to Nyquist's sampling theorem, the sampling rate should be at least twice the highest frequency present in the signal~\citep{nyquist}. If this is not the case, unless effective anti-aliasing techniques are employed, a down-sampled signal can exhibit significant distortions and may appear completely different from the original input signal. 
The aliasing problem exists for most detectors because there are many down-sampling processes involved in CNNs. Down-sampling is a commonly used operation to reduce the number of parameters and discard irrelevant details. Max-pooling, average-pooling, and strided convolution are the most widely applied down-sampling methods in standard CNNs~\citep{resnet, mobilenet, stridedconv}. These down-sampling operations usually ignore Nyquist's sampling theorem, which results in aliasing. The broken object structures and accumulated random noise do harm to the CNN's performance~\citep{aliasing_affect_cnn, wavecnet}.

Previous work has introduced anti-aliased down-sampling operations into CNNs 
~\citep{blurpool,pasa,wavelayer} and has shown the effectiveness of anti-aliasing in tasks such as object classification. However, despite the fact that aliasing is likely to be a particular issue when objects are small, these anti-aliasing techniques have not previously been considered in the field of tiny object detection. This inspires us to design anti-aliased backbones for tiny object detection.
Our proposed anti-aliased backbone is heavily based on WaveCNet~\citep{wavecnet}. WaveCNet employs WaveletPool to replace the down-sampling processes in a standard ResNet~\citep{resnet}, and has been shown to achieve excellent performance in image classification and standard object detection tasks. Here we show that this method can also be used to great effect in tiny object detection.
Our contributions are as follows: 
\begin{itemize}
\item We apply WaveletPool to reduce distortions in the representations of tiny objects and preserve more discriminate information about such objects.
\item We modify WaveCNet so that the order in which Wavelet Pooling is performed is consistent among all connections in the backbone to improve the performance of tiny object detection.
\item We combine anti-aliasing with a bottom-heavy architecture~\citep{bottom_heavy} to improve tiny object detection performance further.
\item We achieve state-of-the-art performance on three tiny object detection benchmarks.
\end{itemize}

\section{Related Work}
\subsection{Tiny Object Detection}\label{sec:tod}
Previous approaches to improving the performance of tiny object detectors can be roughly categorized into seven types \citep{survey}. In the subsequent paragraphs, representative examples from each category will be reviewed.

Super-resolution techniques play a crucial role in enhancing the fine details and characteristics of tiny targets, enabling standard object detection techniques designed for larger objects to be applied effectively to images that contain tiny targets. 
For instance, a generative neural network approach can be employed to enhance the resolution of the entire image~\citep{face_gan}. Alternatively, the super-resolution technique can be selectively applied to small regions of interest within the image~\citep{prior_info_hr}.

Contextual information, obtained from regions of the image surrounding an object, can greatly assist in the detection of that object. Many techniques incorporate contextual cues into their networks. \citet{hr} defined templates that utilized very large receptive fields to capture extensive contextual information to aid in the detection of small faces. Another example is the scale selection pyramid network~\citep[SSPNet;][]{sspnet}. SSPNet encompasses a context attention module (CAM) to generate hierarchical attention heatmaps, allowing the network to prioritize areas that are more likely to contain small objects of interest.

Data augmentation can be used to enhance performance by expanding the training dataset through image transformations. PyramidBox~\citep{pyramidbox} exploited the data-anchor-sampling (DAS) technique that reshaped a randomly selected object within the image to match a smaller anchor size. \citet{tinyperson} proposed an efficient augmentation technique named "scale match" to align the object size distribution differences between general datasets and tiny object detection datasets. The scale match technique involves sampling a size-bin from the histogram of object sizes in the external dataset. To ensure that the range of sampled size-bins does not deviate significantly, the authors also introduced a monotone scale match (MSM) strategy for the sampling in a monotonically increasing or decreasing order. \citet{sm+} introduced an enhanced version of the scale match technique called SM+, which extended it from the image level to the instance level. 

Multi-scale representation learning is a crucial and effective strategy employed in the detection of small or tiny objects. This approach is commonly applied in general object detection to handle objects of varying sizes, such as through the feature pyramid network (FPN) proposed by \citet{fpn}. \citet{fusionfactor} argued that while FPN brought positive impacts, it also introduced negative effects caused by the top-down connections to tiny object detection. To address this, they introduced a statistic-based fusion factor that dynamically adjusted the weight of different layers during feature fusion. \citet{pyramidbox} introduced a low-level feature pyramid network (LFPN) that initiated from a middle layer instead of starting the top-down structure from a high-level layer because of the observation that not all high-level semantic features are equally beneficial for smaller targets. \citet{sfrf} proposed a feature rescaling and fusion (SFRF) network that incorporated a nonparametric adaptive dense perceiving algorithm (NADPA), which had the capability to automatically select and generate a resized feature map that focused on the high-density distribution of tiny objects.

The anchor mechanism predetermines locations which can efficiently scan the image and match potential object regions where objects are anticipated to be detected. \citet{s3fd} maintained consistent anchor density across different scales and proposed a scale compensation anchor matching strategy to ensure that all scales had an adequate number of matched anchors. 
\citet{retinaface} ensured proper matching for objects of all sizes by incorporating anchors of varying sizes across multiple layers. \citet{lsfhi} developed a specialized single-level face detection framework that utilized different dilation rates for anchors with different sizes. \citet{rfla} introduced a novel label assignment method to address the challenges related to insufficient positive samples and the disparity between uniformly distributed prior anchors and the Gaussian-distributed receptive field. To handle the severe misalignment between anchor boxes and axis-aligned
convolutional features, S$^{2}$ANet~\citep{s2anet} included a feature alignment module. In addition to anchor-based methods, some point-based anchor-free methods are also designed for tiny object detection. For example, Oriented RepPoints~\citep{orientedreppoints} represent objects in a set of sample points that effectively bound the spatial extent of the objects and identify semantically significant local areas. 

Training detectors for large objects is relatively straightforward, a significant challenge lies in effectively training detectors for small objects. \citet{sfa} enhanced the robustness and overall performance by using multi-scale training images. Similarly, \citet{hr} proposed an efficient approach where separate face detectors are trained with features extracted from multiple layers of a single feature hierarchy. 

Optimizing the loss function is a valuable strategy that helps enhance the overall performance of detecting tiny objects. \citet{feedback_loss} introduced a novel feedback-driven loss function leveraging the information from the loss distribution as a feedback signal, enabling the model to be trained in a more balanced manner. SSPNet~\citep{sspnet} included an attention-based loss function to supervise the different layers for extracting information from different ranges of objects. 

Our proposed approach considers a new aspect of tiny object detection: anti-aliasing. It is, therefore, independent of the previous work mentioned. Hence, there is a potential that all these different existing techniques could be enhanced by integrating them with our proposed method.

\subsection{Anti-Aliasing Filters}\label{sec:aa}

The low-pass filter is a textbook-style solution to aliasing. Following this idea, several CNN anti-aliasing methods have been proposed~\citep{blurpool, pasa, wavecnet}. \citet{blurpool} proposed BlurPool which incorporated a Gaussian blur layer before each down-sampling module. \citet{bp_order} indicated the order of low-pass filters and convolutions had an impact on the effectiveness of anti-aliasing and updated the residual networks architecture accordingly \citep{bp_order_net}. \citet{pasa} introduced an adaptive content-aware low-pass filtering layer (AdaBlurPool) to generate distinct filter weights for each spatial location and channel group. Although these methods have achieved impressive results on other tasks such as image classification, domain generalization, instance segmentation, and semantic segmentation, they do not seem promising for tiny object detection, and the experimental results in Tab.~\ref{tab:order} confirm this. Gaussian blur~\citep{blurpool, bp_order} is likely to result in a loss of information about tiny features and make tiny objects less recognizable.  Whereas AdaBlurPool~\citep{pasa} is not theoretically justified, and the adaptive low-pass filter is generated by learnable group convolutional layers that are not guaranteed to learn an appropriate low-pass filter (even after many iterations and even if initialized to Gaussian filters).

In contrast to the above methods, \citet{wavecnet} ensures that low-frequency features are not corrupted by high-frequency artifacts by replacing standard down-sampling layers with a discrete wavelet transform (DWT) and an inverse discrete wavelet transform (IDWT). They implemented these transforms using various orthogonal and biorthogonal discrete wavelets including Haar, Daubechies, and Cohen. Such WaveletPool operations were used to replace standard down-sampling operations in ResNets. The resulting 
WaveCNet~\citep{wavecnet} demonstrated the robustness of WaveletPool layers for general object detection in adversarially-attacked scenes. 
Here we apply WaveCNet, with slight modification, to tiny object detection.

\section{Methods}
The impact of aliasing is worse in tiny object detection compared to general object detection because tiny objects have more high-frequency features than general objects. The values of pixels of a tiny object could change rapidly in a small range. Tiny objects are therefore more sensitive to being affected by aliasing. 
However, current backbones used in tiny object detection use CNNs that employ down-sampling operations that ignore Nyquist's sampling theorem, resulting in severe aliasing. To investigate the harm this does to the performance of tiny object detectors we applied WaveletPool (as described in Sec.~\ref{sec:waveletpool}) in a widely used standard backbone ResNet50 (as detailed in Sec.~\ref{sec:waveletnet}) and a bottom-heavy architecture previously shown to be effective for tiny object detection (as described in Sec.~\ref{sec-BHWaveCNet}).

\subsection{Wavelet Pooling}\label{sec:dwt}
\label{sec:waveletpool}

Wavelet Pooling \citep{wavecnet} depends on the wavelet transform theorem~\citep{wavelet}, and is implemented using the DWT and the IDWT. 
\paragraph{1-dimensional (1D) DWT/IDWT.} In 1D, DWT decomposes a given vector $\boldsymbol{x}=\left\{x_{n}\right\}_{n\in \mathbb{Z}}$ into its low-frequency component $\boldsymbol{x}^{\text{low}}=\left\{x_{k}^{\text{low}}\right\}_{k\in \mathbb{Z}}$ and high-frequency component $\boldsymbol{x}^{\text{high}}=\left\{x_{k}^{\text{high}}\right\}_{k\in \mathbb{Z}}$, and IDWT reconstructs the vector $\boldsymbol{x}$ using $\boldsymbol{x}^{low}$ and $\boldsymbol{x}^{high}$, as follows:
\begin{align}
\label{eq:mxlow}&\boldsymbol{x}^{\text {low}}=\mathcal{L} \boldsymbol{x},\quad
\boldsymbol{x}^{\text {high}}=\mathcal{H} \boldsymbol{x}\\
\label{eq:mxrec}&\boldsymbol{x^{*}}=\mathcal{L}^{T} \boldsymbol{x}^{\text{low}} + \mathcal{H}^{T} \boldsymbol{x}^{\text{high}}
\end{align}
where $\boldsymbol{x^{*}}$ denotes the reconstructed vector, $\mathcal{L}=\left\{\boldsymbol{l}_{n-2} ..., \boldsymbol{l}_{n-2k} \right\}^{T}$ and $\mathcal{H}=\left\{\boldsymbol{h}_{n-2}, ..., \boldsymbol{h}_{n-2k} \right\}^{T}$. $\boldsymbol{l}_{j}=\left\{l_{i}\right\}_{i=j-2k,...,j+n-2k}$ and $\boldsymbol{h}_{j}=\left\{h_{i}\right\}_{i=j-2k,...,j+n-2k}$ are the low-pass and high-pass filters of an orthogonal wavelet, $n$ denotes length of the input vector, $k$ denotes the length of the frequency component.

\paragraph{2-dimensional (2D) DWT/IDWT.}
The 2D DWT is achieved by performing 1D DWT on both rows and columns, \ie,
\begin{align}
\boldsymbol{X}_{\text{ll}} =\mathcal{L} \boldsymbol{X} \mathcal{L}^T,\quad
\boldsymbol{X}_{\text{lh}} =\mathcal{H} \boldsymbol{X} \mathcal{L}^T,\quad
\boldsymbol{X}_{\text{hl}} =\mathcal{L} \boldsymbol{X} \mathcal{H}^T,\quad
\boldsymbol{X}_{\text{hh}} =\mathcal{H} \boldsymbol{X} \mathcal{H}^T 
\end{align}
The corresponding 2D IDWT is:
\begin{align}
\boldsymbol{X^{*}} = \mathcal{L}^{T} \boldsymbol{X}_{\text{ll}} \mathcal{L} + \mathcal{H}^{T} \boldsymbol{X}_{\text{lh}} \mathcal{L} + \mathcal{L}^{T} \boldsymbol{X}_{\text{hl}} \mathcal{H} + \mathcal{H}^{T} \boldsymbol{X}_{\text{hh}} \mathcal{H}
\end{align}

Integrating such operations into modern deep learning frameworks is possible as the gradients can be easily calculated~\citep{wavecnet}.

\begin{figure}[tp]
\centering
\includegraphics[width=0.5\linewidth]{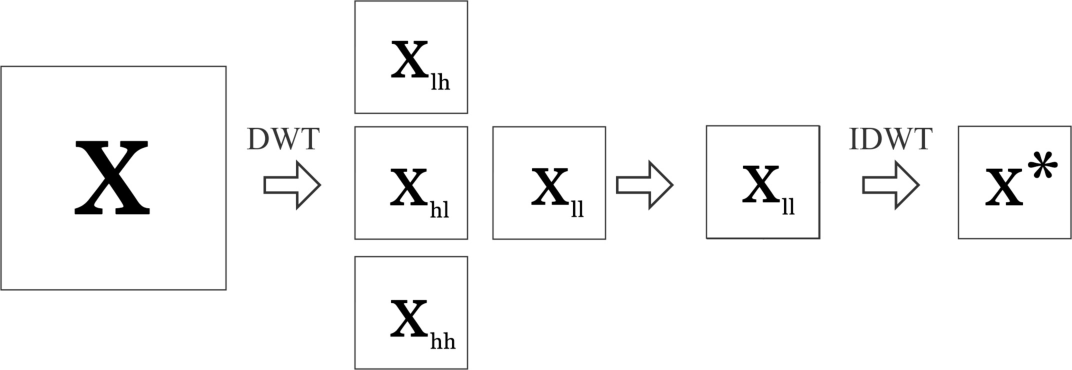}
\caption{Wavelet pooling. The discrete wavelet transformation is applied to the input $\boldsymbol{X}$. The inverse discrete wavelet transformation is applied only to the low-frequency components in order to construct the down-sampled output $\boldsymbol{X^{*}}$. The notations used are the same as those used in the `2-dimensional (2D) DWT/IDWT’ part of Section~\ref{sec:waveletpool}.}\vspace{-5mm}
\label{fig:dwt}
\end{figure}

\paragraph{Down-sampling based on 2D DWT/IDWT} Wavelet Pooling, as illustrated in Fig~\ref{fig:dwt}, consists of applying one 2D DWT and one 2D IDWT in sequence. The application of DWT decomposes the image (or feature maps of a CNN layer) into the high-frequency detail subbands $\boldsymbol{X}_{\text{lh}},\boldsymbol{X}_{\text{hl}},\boldsymbol{X}_{\text{hh}}$, and the low-frequency approximation subband $\boldsymbol{X}_{\text{ll}}$ in the wavelet domain. The IDWT is applied only to low-frequency subband $\boldsymbol{X}_{\text{ll}}$ without using the detail subbands $\boldsymbol{X}_{\text{lh}},\boldsymbol{X}_{\text{hl}},\boldsymbol{X}_{\text{hh}}$, which cuts the resolution in half. The reconstructed image or feature map will still capture the low-frequency information from the input, $\boldsymbol{X}$, while filtering out the high-frequency information without aliasing.

\paragraph{Pooling with different wavelets}
A number of different wavelets can be used. One example of an orthogonal wavelet is the 
Daubechies wavelet. 
It uses $l^{\text{low}}$ and $h^{\text{high}}$ filters to calculate DWT and IDWT. 
If the length of the filter is $2k$, the Daubechies wavelet has an approximation order parameter $k$, noted as Daubechies($k$). Daubechies(1) is also known as Haar wavelet. 

The Cohen wavelet is an example of a biorthogonal wavelet, which means that it uses 
$l^{\text{low}}$ and $h^{\text{high}}$ to implement the DWT but uses their dual vectors $\widetilde l^{\text{low}}$ and $\widetilde h^{\text{high}}$ for the IDWT. A typical Cohen wavelet is written as Cohen($k$, $\widetilde k$) where $k$ and $\widetilde k$ indicate the length of original and dual filters. Cohen(1,1) is identical to the Daubechies(1), which is also the Haar wavelet. In our experiments, we used the Haar wavelet, denoted 'haar', and the Cohen($k$, $\widetilde k$) wavelet, denoted 'ch$k$, $\widetilde k$'.

\subsection{WaveCNet with Consistent Order of Wavelet Pooling}\label{sec:waveletnet}
WaveCNet \citep{wavecnet}  replaces all down-samplings in a ResNet with WaveletPool, as described in the preceding section. Max-pooling is directly replaced by WaveletPool. Strided convolution is replaced by a convolution with the stride of 1 followed by a wavelet-pooling, \ie,
\vspace*{-5mm}\begin{equation}
\begin{split}
MaxPool_{s=2} & \rightarrow WaveletPool_{X_{\text{ll}}} \\
Conv_{s=2} & \rightarrow  Conv_{s=1}\circ WaveletPool_{X_{\text{ll}}}   
\end{split}
\end{equation}
where "Pool" and "Conv" denote pooling and convolution respectively and $\circ$ denotes "followed by".

\begin{figure}[tp]
\centering
\subfigure[]{\includegraphics[width=0.3\linewidth]{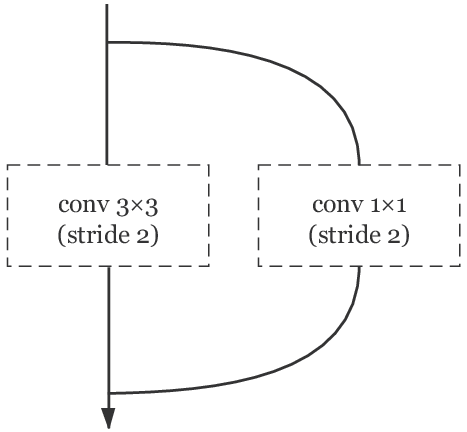}}\hfill
\subfigure[]{\includegraphics[width=0.3\linewidth]{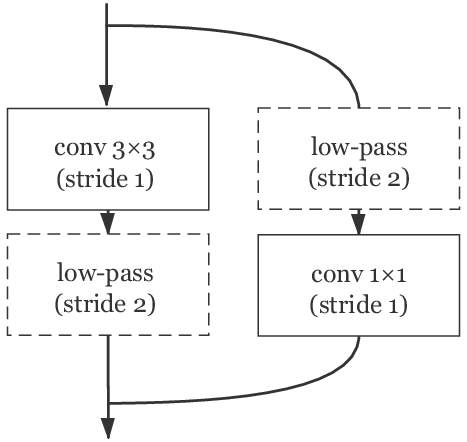}}\hfill
\subfigure[]{\includegraphics[width=0.3\linewidth]{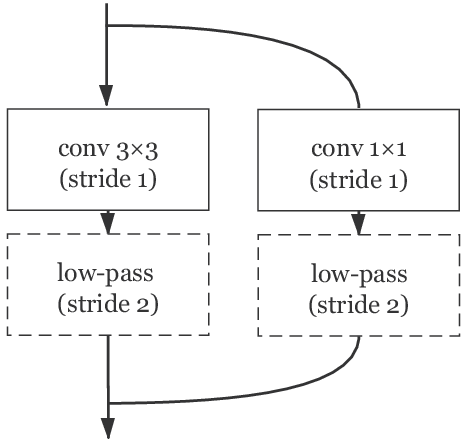}}
\caption{Different implementations of a residual block. (a) original residual block, (b) previous anti-aliasing, (c) our anti-aliasing block.}\vspace{-5mm}
\label{fig:order}
\end{figure}
As discussed in~\citep{bp_order}, whether to apply a low-pass filter before or after a convolutional layer matters for anti-aliasing. Their experiments show better performance is achieved if the low-pass filter comes after the convolution. Despite this, previous anti-aliasing networks~\citep{bp_order, pasa, wavecnet}, have not consistently done so, and the order in which convolution and low-pass filtering is performed differs across different branches in ResNet blocks, as shown in Fig.~\ref{fig:order}.
As Fig.~\ref{fig:order}(a) shows, a standard down-sampling residual block could be simplified as a main path and a skip connection. Fig.~\ref{fig:order}(b) shows the regular order used in most other anti-aliasing CNNs. The low-pass filter follows convolution in the main path but, these operations occur in the reverse order in the skip connection. Experimental results (shown in Tab.~\ref{tab:order}) confirm that using a consistent order in both pathways is beneficial for tiny object detection, and hence, we use the configuration shown in Fig.~\ref{fig:order}(c).
\subsection{Bottom-Heavy Backbone}
\label{sec-BHWaveCNet}

Our previous work showed that using a bottom-heavy version of ResNet50 produced better performance on tiny object detection~\citep{bottom_heavy}. This bottom-heavy backbone, BHResNet50, basically shifts calculations from the top layers to the early layers without introducing additional computational costs. This is done by delaying the down-sampling layers, resulting in a  decrease in the number of convolutional layers applied to low-resolution features in deeper layers, and an increase in the number of convolutional layers used in earlier, higher-resolution, layers. Here, we implement a bottom-heavy anti-aliased architecture by replacing all down-sampling processes with WaveletPool in a BHResNet50.

\section{Experiments}
\subsection{Pre-Training Datasets}
The backbone of an object detector is typically pre-trained on an image classification task as this leads to more rapid and stable training on the detection task~\citep{transfer_learning1, transfer_learning2}. Here, two pre-training datasets were employed.
\paragraph{ImageNet.}
ImageNet, also known as the ImageNet Large Scale Visual Recognition Challenge (ILSVRC) dataset~\citep{in1k}, is a widely used benchmark dataset for image classification tasks in computer vision. It consists of over 1.2 million images from a thousand different categories. The dataset includes a diverse set of categories such as animals, vehicles, household items, plants, and various everyday objects. It is widely recognized as a standard dataset for pre-training the backbone of object detection models. The results shown in Tables~\ref{tab:tinyperson}, \ref{tab:widerface} and \ref{tab:dota} were produced using backbones pre-trained for 100 epochs on ImageNet. The learning rate started from 0.1 and reduced by 1/10 at 30, 60 and 90 epochs.

\paragraph{CIFAR100.}
CIFAR100~\citep{cifar100} is a popular image dataset used for classification tasks in computer vision. It contains 60,000 32x32 colour images. These images are divided into 100 fine-grained classes, with each class having 600 images. The 100 classes in CIFAR100 cover a wide range of object categories, including animals, vehicles, household items, and various natural and man-made objects. CIFAR100 is a good choice for prototyping network architectures because of the reasonable training time. The results shown in Tab.~\ref{tab:order}, \ref{tab:tinyperson_cifar} were obtained using backbones pre-trained for 200 epochs on CIFAR100. The learning rate started from 0.1 and dropped to 1/10 at 100 and 150 epochs.

\subsection{Tiny Object Detection Datasets}
Performance on tiny object detection tasks was evaluated using three standard datasets.
\paragraph{TinyPerson.}
This dataset~\citep{tinyperson} is collected from real-world videos, consisting of 1610 large-scale seaside images with more than 70000 bounding boxes. The 794 images are divided into subsets for training and validation (with 1/10 or the images in the latter). On average, the target objects within the images have an absolute size of approximately 18 pixels. The size range (pixels) of the objects is divided into three sub-intervals for Tinyperson: tiny [2, 20], small [20, 32], and all [2, inf]. The tiny set is partitioned into three (overlapping) sub-intervals: tiny1 [2, 8], tiny2 [8, 12], and tiny3 [12, 20]. Results are reported using mean Average Precision ($mAP$) metrics for all intervals. 
\paragraph{WiderFace.}
This dataset~\citep{widerface} is a widely-used benchmark for face detection. It consists of 32,203 images that contain various human faces with a wide range of scales, poses, and occlusions. The average size of the objects in the WiderFace dataset is approximately 32 pixels. Based on the level of difficulty in detecting the objects, the dataset is divided into three evaluation sets: easy, medium, and hard. Results are reported as $mAP$ for these three subsets. 
\paragraph{DOTA.} This dataset \citep{dota} is for object detection in aerial images. It contains 2,806 aerial images obtained from various sensors and platforms,  with sizes ranging from approximately 800×800 pixels to 4,000×4,000 pixels. The fully annotated DOTA benchmark contains 188,282 instances in 15 common object categories. Half of the original images are randomly assigned to the training set, while 1/6 is allocated to the validation set, and the remaining 1/3 is used as the testing set. We report categorical $AP$ and $mAP$ for the oriented bounding boxes task.

\subsection{Selection of Anti-Aliasing Method and Order of Application}
\tabcolsep3pt
\begin{table}[tp]
\centering
\begin{tabular}{@{}lccccccc@{}}
\toprule[1.5pt]
Anti-Aliasing Method & Order & \textit{$mAP_{tiny}$} & \textit{$mAP_{all}$} & \textit{$mAP_{tiny1}$} & \textit{$mAP_{tiny2}$} & \textit{$mAP_{tiny3}$} & \textit{$mAP_{small}$} \\ \midrule
\multirow{2}{*}{BlurPool}
 & (b) & 41.17 & 43.39 & 24.06 & 44.98 & 54.02 & 58.55\\
 & (c) & 44.65 & 46.74 & 27.71 & 49.45 & 56.14 & 60.30 \\ \midrule
\multirow{2}{*}{AdaBlurPool}
 & (b) & 42.34&45.13&25.21&47.04&54.09&58.30\\
 & (c) & 46.28 & 47.12 & 30.94 & {\bf 52.23} & 56.07 & 59.63 \\ \midrule
\multirow{2}{*}{WaveletPool ch3.3 }
 & (b) &  46.08 & 47.62 & 29.24 & 51.22 & 57.05 & {\bf 63.19}\\
 & (c) & {\bf 46.34} & {\bf 47.70} & {\bf 30.98} & 51.21 & {\bf 57.24} & 60.14 \\ \bottomrule[1.5pt]
\end{tabular}
\caption{Performance of Faster R-CNN on TinyPerson when using a ResNet50 backbone employing different methods of anti-aliasing and different orders of application. (b), (c) indicate the order illustrated in Fig~\ref{fig:order}. Backbones are pre-trained on CIFAR100.}\vspace{-5mm}
\label{tab:order}
\end{table}
As described in Sec.~\ref{sec:aa}, a number of different anti-aliasing methods have been proposed.  Tab.~\ref{tab:order} compares the performance of these different methods on tiny object detection. The results demonstrate that Wavelet Pooling has superior performance, compared to the alternatives, in this application.

As described in Sec.~\ref{sec:waveletnet}, we reverse the order in which convolution and anti-aliased down-sampling are performed in the skip connections in the proposed backbones. The experimental results, shown in Tab.~\ref{tab:order}, show that this change produces a small increase in performance on tiny object detection for Wavelet Pooling and other, alternative, anti-aliased down-sampling methods. We therefore use the order shown in Fig~\ref{fig:order}(c) in all subsequent experiments.

\subsection{Selection of Wavelet}
\tabcolsep2pt
\begin{table}[t]
\centering
\scalebox{0.90}{
\begin{tabular}{@{}llcccccccc@{}}
\toprule[1.5pt]
Back- & Anti-aliasing & \textit{$mAP_{tiny}$} & \textit{$mAP_{all}$} & \textit{$mAP_{tiny1}$} & \textit{$mAP_{tiny2}$} & \textit{$mAP_{tiny3}$} & \textit{$mAP_{small}$} & \textit{Params} & \textit{FLOPs} \\bone \\ \midrule
\multirow{4}{*}{\rotatebox[origin=c]{90}{ResNet}} & None & 44.07 & 45.79 & 28.38 & 49.39 & 55.28 & 60.66 & 41.07 & 75.58 \\
 & WaveletPool haar & 45.37 & 46.62 & 29.78 & 49.26 & 56.51 & 59.70 & 41.30 & 90.78 \\
 & WaveletPool ch3.3 & \textbf{46.34} & 47.70 & \textbf{30.98} & 51.21 & 57.24 & 60.14 & 41.30 & 90.78 \\
 & WaveletPool ch5.5 & 43.63 & 45.43 & 28.96 & 47.24 & 54.84 & 59.87 & 41.30 & 90.78 \\ \midrule
\multirow{4}{*}{\rotatebox[origin=c]{90}{BHResNet}} & None & 44.51 & 45.99 & 30.59 & 50.01 & 54.36 & 60.28 & 28.75 & 75.79 \\
 & WaveletPool haar & 45.33 & 46.95 & 30.15 & 49.48 & 56.72 & \textbf{61.54} & 28.75 & 91.90 \\
 & WaveletPool ch3.3 & 46.04 & \textbf{47.70} & 29.28 & \textbf{51.21} & \textbf{57.33} & 61.01 & 28.75 & 91.90 \\
 & WaveletPool ch5.5 & 44.61 & 46.18 & 29.34 & 48.97 & 55.18 & 60.15 & 28.75 & 91.90 \\ \bottomrule[1.5pt]
\end{tabular}}
\caption{Performance of Faster R-CNN on TinyPerson when using different anti-aliasing filters. Backbones are pre-trained on CIFAR100. `Params' indicates the number of parameters (in millions) required by the detector. `FLOPs' indicates the number of floating-point operations (in billions) performed on the standard TinyPerson input size of 640×512 pixels. Wavelets compared in this table are described in the `Pooling with different wavelets' part of Section~\ref{sec:waveletpool}.}
\label{tab:tinyperson_cifar}
\end{table}
To select an appropriate wavelet to use in the anti-aliasing module, the performance produced with different wavelets was tested on TinyPerson. These results were compared to those produced using a standard CNN backbone. These backbones were pre-trained on CIFAR100 to reduce pre-training time. As shown in Tab.~\ref{tab:tinyperson_cifar}, Cohen(3,3) wavelet produced the best results for most metrics, both for the ResNet50 and BHResNet50 backbones. Cohen(3,3) was, therefore, selected as the wavelet used in subsequent experiments. 

\subsection{Results on TinyPerson}
\begin{table}[tp]
\tabcolsep1.5pt
\hspace*{-1mm}\scalebox{0.90}{
\begin{tabular}{@{}lcccccc@{}}
\toprule
Methods & \textit{$mAP_{tiny}$} & \textit{$mAP_{all}$} & \textit{$mAP_{tiny1}$} & \textit{$mAP_{tiny2}$} & \textit{$mAP_{tiny3}$} & \textit{$mAP_{small}$} \\ \midrule
Faster R-CNN~\citep{faster}&47.81&49.99&31.78&53.54&58.34&64.56\\ 
Adap FCOS~\citep{fcos}&48.45&51.03&29.75&52.71&60.36&65.28\\
S-$\alpha$ + Faster R-CNN~\citep{fusionfactor}&48.39&*&31.68&52.20&60.01&65.15\\
RFLA + Faster R-CNN~\citep{rfla}&48.86&51.92&30.35&54.15&61.28&66.69\\
SM + Faster R-CNN~\citep{tinyperson}&49.09&51.57&32.93&55.42&59.70&64.24\\
SM+ + Faster R-CNN~\citep{sm+}&51.46&*&33.74&55.32&62.95&67.37\\
SFRF + Faster R-CNN~\citep{sfrf}&57.24&59.03&51.49&64.51&67.78&65.33\\ 
SSPNet w ResNet50~\citep{sspnet} & 57.93 & 62.23 & 45.33 & 60.26 & 67.09 & 71.56\\ 
SSPNet w BHResNet50~\citep{bottom_heavy} & 58.97 & 62.02 & 47.22 & 61.61 & 67.45 & 72.37\\ \midrule
SSPNet w ResNet50 + WaveletPool & 59.24 & 63.16 & \textbf{47.76} & 60.23 & 68.04 & 73.42\\
SSPNet w BHResNet50 + WaveletPool & \textbf{59.34} & \textbf{64.02} & 46.55 & \textbf{61.73} & \textbf{68.95} & \textbf{73.87}\\ \bottomrule
\end{tabular}}
\caption{Comparison of the performance of the proposed methods with previous results on TinyPerson. All backbones were pre-trained on ImageNet. * indicates results that are unavailable because of unreleased code.}\vspace{-5mm}
\label{tab:tinyperson}
\end{table}
By replacing the standard ResNet50 backbone in the current state-of-the-art method SSPNet with our proposed anti-aliasing backbones,
we get better $mAP$ among all intervals of TinyPerson. As shown in Tab.~\ref{tab:tinyperson}, using WaveletPool increases the $mAP_{tiny}$ by 1.31\% compared to plain SSPNet while integrating WaveletPool with the bottom-heavy backbone increases performance by an additional 0.1\%. Hence, overall the proposed method improves on the previous state-of-the-art result by 1.41\% in terms of $mAP_{tiny}$. All networks were trained for 12 epochs with stochastic gradient descent (SGD) as the optimizer. The learning rate was initialized to 0.002/0.003 for ResNet50+WaveletPool and BHResNet50+WaveletPool respectively and decreased by a factor of 0.1 after 8 and 11 epochs. 

\subsection{Results on WiderFace}

\begin{table}[tp]
\tabcolsep3.5pt
\centering
\begin{tabular}{@{}lccc@{}}\toprule
Method & $mAP_{easy}$ & $mAP_{medium}$ & $mAP_{hard}$ \\ \midrule
HR~\citep{hr}&92.5&91.0&80.6\\
S3FD~\citep{s3fd}&93.7&92.4&85.2\\
FaceGAN~\citep{face_gan}&94.4&93.3&87.3\\
SFA~\citep{sfa}&94.9&93.6&86.6\\
LSFHI~\citep{lsfhi}&95.7&94.9&89.7\\
Pyramid-Box~\citep{pyramidbox}&96.1&95.0&88.9\\
RetinaFace~\citep{retinaface}&96.5&95.6&90.4\\
TinaFace w ResNet50~\citep{tinaface}&96.3&95.7&93.1\\ \midrule
TinaFace w ResNet50 + WaveletPool &96.4&95.8&93.1\\
TinaFace w BHResNet50 + WaveletPool &\textbf{96.6}&\textbf{96.0}&\textbf{93.4}\\ \bottomrule
\end{tabular}
\caption{Performance of previous methods and our methods on WiderFace.}
\label{tab:widerface}
\end{table}

Table~\ref{tab:widerface} shows that integrating WaveletPool into the ResNet50 backbone, and its bottom-heavy variant, used by TinaFace~\citep{tinaface}, the current state-of-the-art method for WiderFace, achieves new state-of-the-art results on all three sub-tasks. Both backbones are trained using SGD optimizer with batch size 16. Training was performed for a total of 630 epochs and the learning rate varied between 3.75$e$-3 and 3.75$e$-5 every 30 epochs following the cosine schedule~\citep{cosdecay}. 

\subsection{Results on DOTA}
\tabcolsep0.8pt
\begin{table}[tp]
\begin{center}
\scalebox{0.8}{
\begin{tabular}{@{}lcccccccccccccccc@{}}
\toprule[1.5pt]
Method & PL & BD & BR & GTF & SV & LV & SH & TC & BC & ST & SF & RO & HA & SP & HE & $mAP$ \\ \midrule
SSD~\citep{ssd} & 39.8 & 9.1 & 0.6 & 13.2 & 0.3 & 0.4 & 1.1 & 16.2 & 27.6 & 9.2 & 27.2 & 9.1 & 3.0 & 1.0 & 1.0 & 10.6 \\
FDLoss~\citep{feedback_loss} & 79.0 & 38.2 & 28.7 & 36.9 & 44.2 & 40.9 & 57.8 & 65.4 & 54.6 & 36.3 & 34.1 & 39.6 & 42.9 & 45.1 & 16.3 & 43.2 \\
S$^{2}$ANet~\citep{s2anet} & \textbf{89.2} & \textbf{78.0} & 47.7 & 68.1 & 77.4 & 73.1 & 79.1 & 90.9 & 79.5 & 85.0 & 56.7 & 58.6 & 59.9 & 65.6 & \textbf{51.2} & 70.1 \\
RepPoints~\citep{reppoints} & 87.8 & 77.7 & 49.5 & 66.5 & 78.5 & 73.1 & 86.6 & 90.7 & \textbf{83.8} & 84.3 & 53.1 & \textbf{65.6} & 63.7 & \textbf{68.7} & 45.9 & 71.7 \\ \midrule
RepPoints+WaveletPool & 88.6 & 73.8 & \textbf{52.4} & 70.9 & 79.1 & 73.3 & 86.2 & \textbf{90.9} & 81.2 & \textbf{85.7} & 58.8 & 61.4 & \textbf{64.5} & 66.2 & 45.4 & \textbf{71.9} \\
RepPoints w BH+WaveletPool & 88.3 & 74.6 & 49.8 & \textbf{72.4} & \textbf{79.4} & \textbf{74.0} & \textbf{86.6} & 90.8 & 78.6 & 84.7 & \textbf{59.9} & 60.1 & 62.6 & 67.0 & 44.7 & 71.6 \\ 
\bottomrule[1.5pt]
\end{tabular}}
\end{center}
\caption{Categorical and mean average precision of previous and our methods on DOTA. BH indicates BHResNet50. The abbreviations indicate the following categories: PL-Plane, BD-Baseball diamond, BR-Bridge, GTF-Ground field track, SV-Small vehicle, LV-Large vehicle,
SH-Ship, TC-Tennis court, BC-Basketball court, ST-Storage tank, SBF-Soccer-ball field, RA-Roundabout, HA-Harbor, SP-Swimming pool, and HC-Helicopter.}\vspace{-5mm}
\label{tab:dota}
\end{table}
For the DOTA detection task, we evaluated the proposed backbones incorporating WaveletPool by integrating them with the current state-of-the-art method: Oriented RepPoints \citep{orientedreppoints}. The proposed method achieved the best $mAP$, and better $AP$ for the majority of object categories. Both backbones were trained for 12 epochs with SGD as the optimizer. The learning rate was initialized to 0.008 and decreased by a factor of 0.1 after 8 and 11 epochs. The results were produced using the DOTA evaluation server.


\subsection{Knowledge-Distillation for Efficient Learning}
Knowledge distillation is a method used in machine learning to transfer knowledge from an accurate model, known as the teacher model, to a simpler model, called the student model~\citep{kd}. The goal is to distill the knowledge learned by the teacher model into the student model, enabling the student model to achieve comparable performance to the teacher model but with reduced computational requirements. Inspired by~\citep{kt}, we use the original ResNet50 as the teacher and an architecture incorporating WaveletPool (either ResNet50 or BHResNet50) as the student. The student network was trained using a short training schedule which had only ten epochs and the learning rate was reduced by 1/10 at epochs 3, 6 and 9. As can be seen in Tab.~\ref{tab:kd}, using a short schedule alone clearly has a negative impact on the performance of the model. However, when combined with the knowledge-distillation approach, the achieved $mAP$ was similar to that achieved using the original, full, training schedule. This demonstrates that it is possible to train the proposed backbones efficiently which could be helpful for exploring different architectures based on WaveletPool in future research. 

\tabcolsep5pt
\begin{table}[tp]
\centering
\begin{tabular}{@{}lllcccr@{}}
\toprule
Methods & SS & KD & $mAP_{tiny}$ & $mAP_{all}$ & Top-1 Acc & Training Time \\ \midrule
ResNet50 & & & 51.72 & 53.62 & 77.47 & 3807 mins \\
 + WaveletPool&\Checkmark &  & 49.52 & 51.12 & 65.31 & 284 mins \\
&\Checkmark & \Checkmark & 51.42 & 53.66 & 73.05 & 283 mins \\ \midrule
BHResNet50  & & & 51.78 & 53.60 & 76.31 & 3807 mins \\
 + WaveletPool&\Checkmark &  & 49.37 & 51.51 & 64.51 & 284 mins \\
&\Checkmark & \Checkmark & 50.61 & 53.39 & 70.86
 & 283 mins \\
\bottomrule
\end{tabular}
\caption{Performance of the proposed backbones when trained using different methods and schedules on TinyPerson. All results are evaluated with Faster R-CNN. SS indicates the short schedule. KD indicates the knowledge distillation training method.}\vspace{-5mm}
\label{tab:kd}
\end{table}

\section{Conclusion}
This study explores the detrimental impact of aliasing on tiny object detection. To suppress the aliasing effect, we replace standard down-sampling methods in the commonly used ResNet50 backbone with WaveletPool. Experimental results demonstrate the superior object detection accuracy achieved by our approach. We have achieved state-of-the-the art performance on DOTA using this method. Furthermore, extending our previously proposed bottom-heavy backbone architecture to also use WaveletPool, we achieve state-of-the-art performance on the TinyPerson and WiderFace datasets while utilizing fewer parameters. Additionally, we explore the potential for expediting the training process of our network using knowledge distillation techniques, providing an efficient route for future investigations to incorporate our methods into other backbones.

\acks{The authors gratefully acknowledge the use of the King's Computational Research, Engineering and Technology Environment (CREATE) and Joint Academic Data Science Endeavour (JADE). This research was funded by the King's-China Scholarship Council (K-CSC).}

\bibliography{acml23}

\appendix





\end{document}